\documentclass[runningheads]{llncs}
\usepackage[utf8]{inputenc}
\usepackage{graphicx}
\usepackage[table,dvipsnames]{xcolor}
\usepackage{amsmath}
\usepackage{amssymb}
\usepackage{bm}
\usepackage{gensymb}
\usepackage{siunitx}
\usepackage{microtype}
\usepackage{xspace}
\usepackage{siunitx}
\usepackage[super]{nth}
\usepackage[nolist]{acronym}
\usepackage[normalem]{ulem}
\usepackage[hyphens]{url}
\usepackage{array}
\usepackage{multirow}
\usepackage[export]{adjustbox}
\usepackage{tabulary}
\usepackage{collcell}
\usepackage{hhline}

\usepackage{booktabs}
\usepackage[pagebackref=false,%
	breaklinks=true,colorlinks,bookmarks=false,citecolor=green!80!black,linkcolor=red!80!black,urlcolor=blue]{hyperref}
\usepackage{cleveref}
\usepackage{subcaption}
\usepackage{tikz}
\usetikzlibrary{%
shadows,spy,decorations,plotmarks,matrix,mindmap,%
arrows,arrows.meta,patterns,external,pgfplots.groupplots,fit,calc,positioning,shapes,
backgrounds,decorations.markings}
\usepackage{pgfplots}
\pgfplotsset{compat=newest}

\crefname{section}{Sec.}{Sections}
\crefname{figure}{Fig.}{Figure}
\crefname{table}{Tab.}{Table}
\crefname{equation}{Equ.}{Equation}

\usepackage[inline]{enumitem}
\setlist*[enumerate]{label=(\arabic*)}

\begin{acronym}
	\acro{cnn}[CNN]{Convolutional Neural Network}
	\acro{dl}[DL]{Deep Learning}
	\acro{gmp}[GMP]{Generalized Max Pooling}
	\acro{ocr}[OCR]{Optical Character Recognition}
	\acro{nn}[NN]{Neural Network}
	\acro{rnn}[RNN]{Recurrent Neural Network}
	\acro{sift}[SIFT]{Scale Invariant Feature Transform}
	\acro{svm}[SVM]{Support Vector Machine}
	\acro{vlad}[VLAD]{Vectors of Locally Aggregated Descriptors}
\end{acronym}

\newcommand{\onedot}{.\xspace}
\newcommand{\etal}[1]{#1~et~al\onedot}

\newcommand{\eg}{e.\,g.,\xspace}

\newcommand{\cf}{cf\onedot}
\newcommand{\ie}{i.\,e.,\xspace}

\tikzstyle{blockbody} = [rectangle, draw, fill=blue!20, 
		text centered, rounded corners, 
	minimum height=2em, thick]
\tikzstyle{blockwhite} = [rectangle, draw, 
		text centered, rounded corners, 
	minimum height=2em, thick]
\tikzstyle{block} = [rectangle, draw=black, fill=blue!20,
		text width=5.5em, text centered, rounded corners, 
	minimum height=2em, thick]
\tikzstyle{blockzero} = [rectangle, draw=white, fill=white,
		text=white, text width=5.5em, 
		text centered, rounded corners, 
	minimum height=2em, thick]

\tikzstyle{blockh} = [rectangle, draw, fill=red!20, 
		text width=5.5em, text centered, rounded corners, 
	minimum height=2em, thick]
	
\tikzstyle{blockl} = [rectangle, draw, fill=blue!10, 
		text width=5.5em, text centered, rounded corners, 
	minimum height=2em, thick]
\tikzstyle{blocklgray} = [rectangle, draw, fill=blue!10, text=gray!70,
		text width=5.5em, text centered, rounded corners, 
	minimum height=2em, thick]
	\tikzstyle{blocklnotext} = [rectangle, draw, fill=blue!10, text=blue!10,
		text width=5.5em, text centered, rounded corners, 
	minimum height=2em, thick]

\tikzstyle{roundish} = [draw, rectangle, fill=red!20, rounded corners,
text width=5em, text centered, minimum height=2em, thick, inner sep=0em]
\tikzstyle{hhilit} = [draw=black, thick, dotted, inner xsep=0.5em, 
inner	ysep=0.5em]
\tikzstyle{newpart}=[align=center,black,thin,draw, rounded	corners,
fill=green!20]
\tikzstyle{newerpart}=[align=center,black,thin,draw, rounded	corners,
fill=orange!20]

\tikzset{>=stealth'}

\def\colorModel{hsb} %
\newcommand\ColCell[1]{
  \pgfmathparse{#1<50?1:0}  %
    \ifnum\pgfmathresult=0\relax\color{white}\fi
  \pgfmathsetmacro\compA{0}      %
  \pgfmathsetmacro\compB{#1/100} %
  \pgfmathsetmacro\compC{1}      %
  \edef\x{\noexpand\centering\noexpand\cellcolor[\colorModel]{\compA,\compB,\compC}}\x #1
  } 

\newcolumntype{E}{>{\collectcell\ColCell}m{0.2cm}<{\endcollectcell}}  %
\NewDocumentCommand{\rot}{O{60} O{1em} m}{\makebox[#2][l]{\rotatebox{#1}{#3}}}%
\NewDocumentCommand{\rotn}{O{90} O{1em} m}{\makebox[#2][l]{\rotatebox{#1}{#3}}}%

\begin{document}
\title{Writer Retrieval and Writer Identification in Greek Papyri}
\author{Vincent~Christlein*\orcidID{0000-0003-0455-3799} \and
Isabelle~Marthot-Santaniello$^\star$\orcidID{0000-0003-0407-8748} \and
Martin~Mayr*\orcidID{0000-0002-3706-285X} \and
Anguelos~Nicolaou*\orcidID{0000-0003-3818-8718} \and
Mathias~Seuret*\orcidID{0000-0001-9153-1031}
}
\authorrunning{V.\ Christlein et al.}
\institute{*Pattern Recognition Lab, Friedrich-Alexander-Universität Erlangen-Nürnberg, Erlangen, Germany\\
$^\star$Departement Altertumswissenschaften, Universität Basel, Basel, Switzerland
\email{vincent.christlein@fau.de}}
\maketitle              %
\begin{abstract}
The analysis of digitized historical manuscripts is typically addressed by paleographic experts. 
Writer identification refers to the classification of known writers while writer retrieval seeks to find the writer by means of image similarity in a dataset of images. 
While automatic writer identification/retrieval methods already provide promising results for many historical document types, papyri data is very challenging due to the fiber structures and severe artifacts. 
Thus, an important step for an improved writer identification is the preprocessing and feature sampling process. 
We investigate several methods and show that a good binarization is key to an improved writer identification in papyri writings.
We focus mainly on writer retrieval using unsupervised feature methods based on traditional or self-supervised-based methods. 
It is, however, also comparable to the state of the art supervised deep learning-based method in the case of writer classification/re-identification. 
\end{abstract}
\section{Introduction}
The mass digitization of handwritten documents not only makes them accessible to the public, but also accelerates research in the fields of linguistics, history and especially paleography. 

An important task is writer\footnote{``Writer'' and ``scribe'' is used interchangeably throughout the paper.} identification (scribe attribution), which can provide clues to life in the past and enable further analysis of networks, sizes of writing schools, correspondences, etc. 
The term writer identification is often used for both \emph{writer retrieval} and \emph{writer classification} (or writer re-identification). 
Writer retrieval is related to the scenario where a query image with possible known writer identity is given and a dataset of images is ranked according to their similarity to the query image. 
This can help to pre-sort large corpora. 
Conversely, writer re-identification has a training set of known identities that can be used to train classifiers that are able to distinguish and classify new samples in these classes. 

Writer identification can be obtained at two levels: based on the content of the text or on the appearance of the writing. 
For the first one, the raw text, \ie a transcription of the text, is analyzed on the stylistic point of view and the author is attributed. 
However, the author does not need to be the writer of the text, which could for example be written by a secretary. 
An appearance analysis can give clues about the person who penned the text, the actual writer. 
This appearance analysis can be realized in multiple ways.
Sometimes the layout can give hints about the scribe, \eg in charters~\cite[pp.\,40]{Krafft10}.
However, this depends much on the document type and its tradition quality.
Writer identification is traditionally a completely exemplar-based and manual task often accomplished by forensic or paleographic experts.
Results obtained by paleographers have often been put in doubt for their subjectivity, the difficulty to produce objective, reproducible arguments.   
In this paper, we explore an automatic writer identification based upon the script appearance using signed documents, \ie internal-based evidence of the writer's identity. 
We enrich our interpretation of the results with paleographic considerations in order to better apprehend the peculiarities of this group of writers.

Script-based writer identification in historical documents gained some attention throughout the last years. 
A popular group of approaches analyses the textural components of script, \eg angles or script width, such as the \emph{Hinge} feature~\cite{Bulacu03}, which measures at each contour point two different angles. 
This feature was also applied~\cite{Bulacu07AHI} to a medieval dataset containing \num{70} document images written by ten writers.
Another textural feature is the \emph{Quill} feature~\cite{Brink12WIU}.
It relates the ink width to the angle at a specific radius.
The authors show promising results on historical data consisting of \num{10}--\num{18} writers (depending on the language) using \num{70}--\num{248} document images.

Fiel and Sablating~\cite{Fiel14His} suggest a method based on \ac{sift} and Fisher Vectors~\cite{Fiel13WIW} to historical Glagolitic documents (oldest known Slavic alphabet) consisting of \num{361} images written by eight scribes.
Therefore, they detect the text by local projection profiles and binarize the document images using the method of \etal{Su}~\cite{Su10}.

Another group of scientists investigate methods for writer identification in
historical Arabic documents~\cite{Fecker14DWA,Fecker14WIF,Asi17}. 
The rejection strategy proposed by \etal{Fecker}~\cite{Fecker14DWA} is especially interesting, since it gives information whether a writer is present in the reference dataset or not.

We investigated writer identification for different letter sets of correspondences~\cite{Christlein16AWI}. 
We also proposed a deep learning-based approach based on self-supervised learning. 
It still achieves close to state-of-the-art results on the dataset of letters of the ICDAR'17 competition on historical writer identification~\cite{Fiel17ICDAR}. 

Apart from the technical point of view, writer identification methods were also used in various paleographic analyses. 
The effort of finding join candidates of the Cairo Genizah collection, which consists of approximately 350 000 fragments, could be reduced drastically by an automatic method for finding join candidates focusing on similar writing~\cite{Wolf11}.
More recently, \etal{Shaus}~\cite{Shaus20} analyzed 18 Arad inscriptions (ca.\ 600 BCE). 
By means of a forensic expert and statistical methods, they show that the analyzed corpus was written by at least 12 different writers. 
\etal{Popović}~\cite{Popovic21} studied the Bible's ancient scribal culture by analyzing the Great Isaiah Scroll (1QIsa$^a$) of the Dead Sea Scrolls and found a switching point in the column series with a clear phase transition in several columns. 
This suggests that two main scribes were writing this scroll, showing that multiple scribes worked at the same manuscript. 

Papyri manuscripts were studied by \etal{Pirrone}~\cite{Pirrone21} who investigate self-supervised deep metric learning for ancient papyrus fragments retrieval. 
Papyri data was also used as a challenging dataset in a competition for segmenting the text (binarization)~\cite{Pratikakis19}. 
A baseline for writer identification of papyri manuscripts was given by \etal{Mohammed}~\cite{Mohammed19}. 
It is based on \ac{sift} descriptors evaluated on FAST keypoints classified using a normalized local naive Bayes Nearest Neighbor classifier.
The authors also provided a first papyri dataset (GRK-Papyri) consisting of 50 image samples of 10 writers. 
\etal{Cilia}~\cite{Cilia21} used an extended version of this dataset (122 images from 23 writers) and extracted single lines to form the PapyRow dataset that can be used for further investigation.  
However note that results based on lines are not image-independent anymore and thus could be biased by background artifacts. 
The GRK-Papyri dataset was also in focus of two further studies~\cite{Nasir21LFF,Nasir21WCF} that focused on writer (re-)identification, \ie a part of the dataset was used for training and the task was to classify the remainder of the dataset. 
They showed that a two-step fine-tuning is beneficial for writer identification. 
In particular, an ImageNet-pretrained network was fine-tuned first on the IAM writer identification dataset~\cite{Marti02IAM} before it was fine-tuned a second time on the writers of the papyri training dataset. 

In this work, we evaluate two methods for writer identification. 
Both methods are completely unsupervised, \ie they do not need known writer identities. 
One method relies on traditional features and the other one trains a deep neural network in a self-supervised fashion. 
We evaluate the methods on the GRK-Papyri dataset~\cite{Mohammed19} where we mainly focus on the retrieval scenario because there are no other works apart of a baseline given by \etal{Mohammed}~\cite{Mohammed19} considering the retrieval case. 
We show that both methods surpass the baseline by a large margin. 
In particular, we investigate the critical sampling procedure and show that a good binarization increases the retrieval and re-identification accuracy. 

The paper is organized as follows. 
We first present the methodology used for writer identification in \cref{sec:methodology}. 
In \cref{sec:evaluation}, we first discuss the dataset, then present two different evaluation protocols: writer retrieval and writer classification/re-identification, and eventually present and discuss our results before we conclude the paper in \cref{sec:conclusion}.

\section{Methodology} \label{sec:methodology}
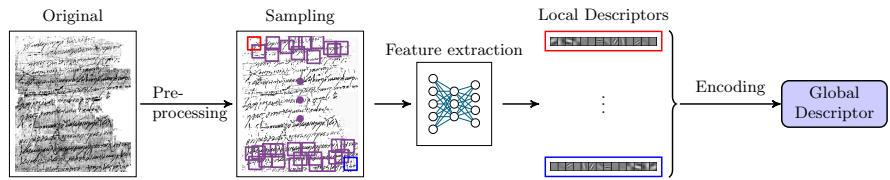
\begin{figure}[t]
	\centering
	\resizebox{0.97\textwidth}{!}{
	\begin{tikzpicture}%

		\node[](nnet){
		\includegraphics[width=0.08\textwidth]{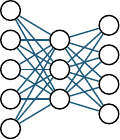}
		};

		\node[fit=(nnet),draw](fit){};
		\node[above=0.01cm of fit]{Feature extraction};

		\node[rectangle,draw,left=1cm of fit](imgbin){
			\includegraphics[width=0.18\textwidth]{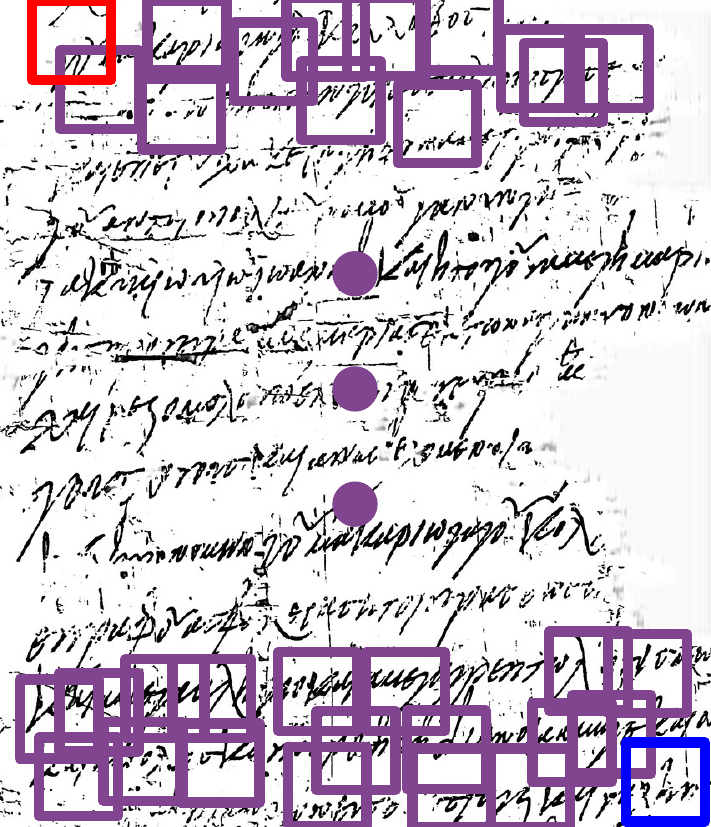} %
		};
		\node[above=0.01cm of imgbin]{Sampling};

	\draw[->, thick]($(imgbin.east)+(0.2cm,0)$)--($(fit.west)+(-0.1cm,0)$);
	
	\node[rectangle,draw,left=1.9cm of imgbin](img){
			\includegraphics[width=0.18\textwidth]{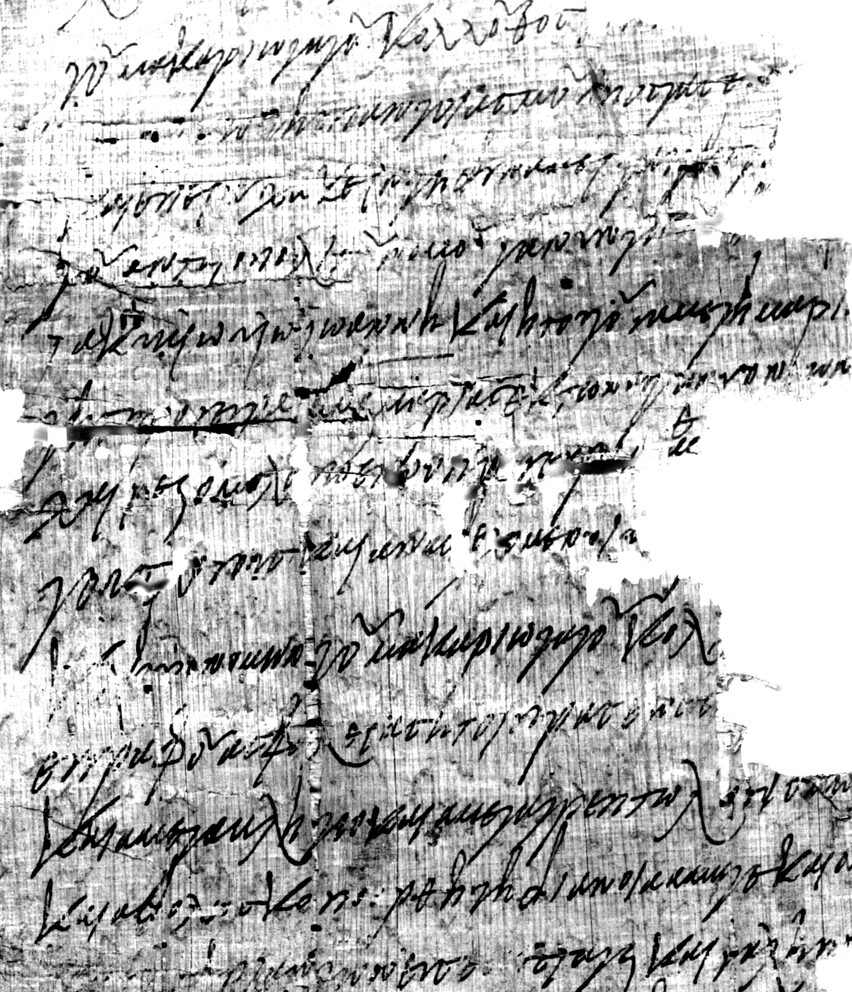} %
		};
		\node[above=0.01cm of img]{Original};

		\draw[->, thick,shorten >=2pt, shorten <=2pt](img) --
		(imgbin)node[midway,sloped,text width=1.3cm]{Pre-\\processing};

			\node[right=1cm of fit,draw,red,thick,anchor=south
			west,yshift=1cm](af){
				\includegraphics[width=2cm]{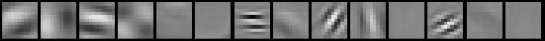}
			};
			\coordinate(c1) at (af.west |- fit);
	\draw[->, thick]($(fit.east)+(0.2cm,0)$)--($(c1.west)+(-0.1cm,0)$);
	
	\node[right=1cm of fit,anchor=north west,
	draw,blue,thick,yshift=-1cm](af2){
				\includegraphics[width=2cm]{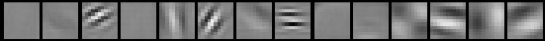}
			};
			\node[yshift=0.1cm] (dots) at ($(af)!0.5!(af2)$) {$\vdots$};
	\node[above=0.02cm of af] {Local	Descriptors};
	
	\draw [decorate,decoration={brace,amplitude=4pt,raise=2pt},thick,draw=black]
	($(af.north east) + (0.05cm,0)$) -- ($(af2.south east) + (0.05cm,0)$)
	node [%
	midway,align=center,sloped,above=1pt](b){};
		\node[block, right=5.5cm of fit, text width=5.5em, minimum height=2em] (sv) {Global Descriptor};
		\draw[->,thick](b) -- (sv)
		node[midway,sloped,above=1pt,rounded corners](agg){Encoding};
	\end{tikzpicture}
	} %
  \caption{Writer identification pipeline.}
  \label{fig:pipeline}
\end{figure}

The writer identification methods we use are based on the work done for the PhD thesis of V.\ Christlein~\cite{Christlein19PHD}.  
It follows a writer identification pipeline based on local features, see \cref{fig:pipeline}. 
First, the image is pre-processed. 
This typically involves the binarization of the image. 
Later, we will see that this is a particular important part of the pipeline. 
We extract local features at specific sampling points in the image, \eg from contours of the handwriting or from keypoints. 
The local descriptors themselves could be generated by a neural network as depicted in \cref{fig:pipeline}. 
For example in a supervised way by training \ac{cnn}~\cite{Christlein15GCPR,Fiel15CAIP}. 
However, we do not have much training data available. 
Therefore, we rely on two unsupervised methods to create local descriptors:
\begin{enumerate}
    \item We use descriptors based on \ac{sift}~\cite{Lowe04DIF}. The \ac{sift} descriptors are Dirichlet-normalized, decorrelated and dimensionality-reduced to 64 components by means of PCA-whitening, and eventually $\ell^2$-normalized, \ie the representation is normalized such that its $\ell^2$ norm is one. More details are given in~\cite{Christlein19PHD}.

\item We learn features in a self-supervised fashion~\cite{Christlein17ICDAR} using a \ac{cnn}, \ie without the need of labeled training data (here the writer information). 
At \ac{sift} keypoints, \ac{sift} descriptors and \numproduct{32 x 32} patches are extracted.
This patch size was experimentally shown to work best.
The \ac{sift} descriptors are clustered using $k$-means. 
The patches are used as input for a \ac{cnn}. 
As targets for the \ac{cnn}, we use the cluster ids of the \ac{sift} clusters.
Note that we omit \ac{sift} keypoints and descriptors corresponding to patches that are completely blank.
Furthermore, because the image resolution is mostly very large, we downsample the images by a factor of two in each dimension. 
\end{enumerate}

From these local descriptors, we create global descriptors to enable a simplified matching.
We rely on \ac{vlad}~\cite{Jegou12ALI} for encoding the local descriptors into a high dimensional global representation using 100 clusters for $k$-means clustering.
Additionally, we employ \ac{gmp}, which was shown to improve the writer identification performance~\cite{Christlein18DAS,Christlein19ICDAR} with a regularization parameter $\gamma=1000$.
The global representation is eventually power-normalized with a power of $0.5$ and $\ell^2$-normalized. 
This process is repeated five times. 
These five representations are afterwards jointly PCA-whitened and once more $\ell^2$-normalized to further decorrelate the global representations~\cite{Christlein19PHD}.
This final representation is eventually used for similarity comparison using Cosine distance, which is a common metric for writer retrieval. If a large enough independent training set would be available, then also Exemplar-SVMs could be used to improve the results further~\cite{Christlein17PR}. 

\section{Evaluation} \label{sec:evaluation}
\subsection{Data}
\label{sec:data}
\begin{figure}[t]
\centering
	\begin{tikzpicture}
		\node[](i1){
			\includegraphics[height=0.3\textheight]{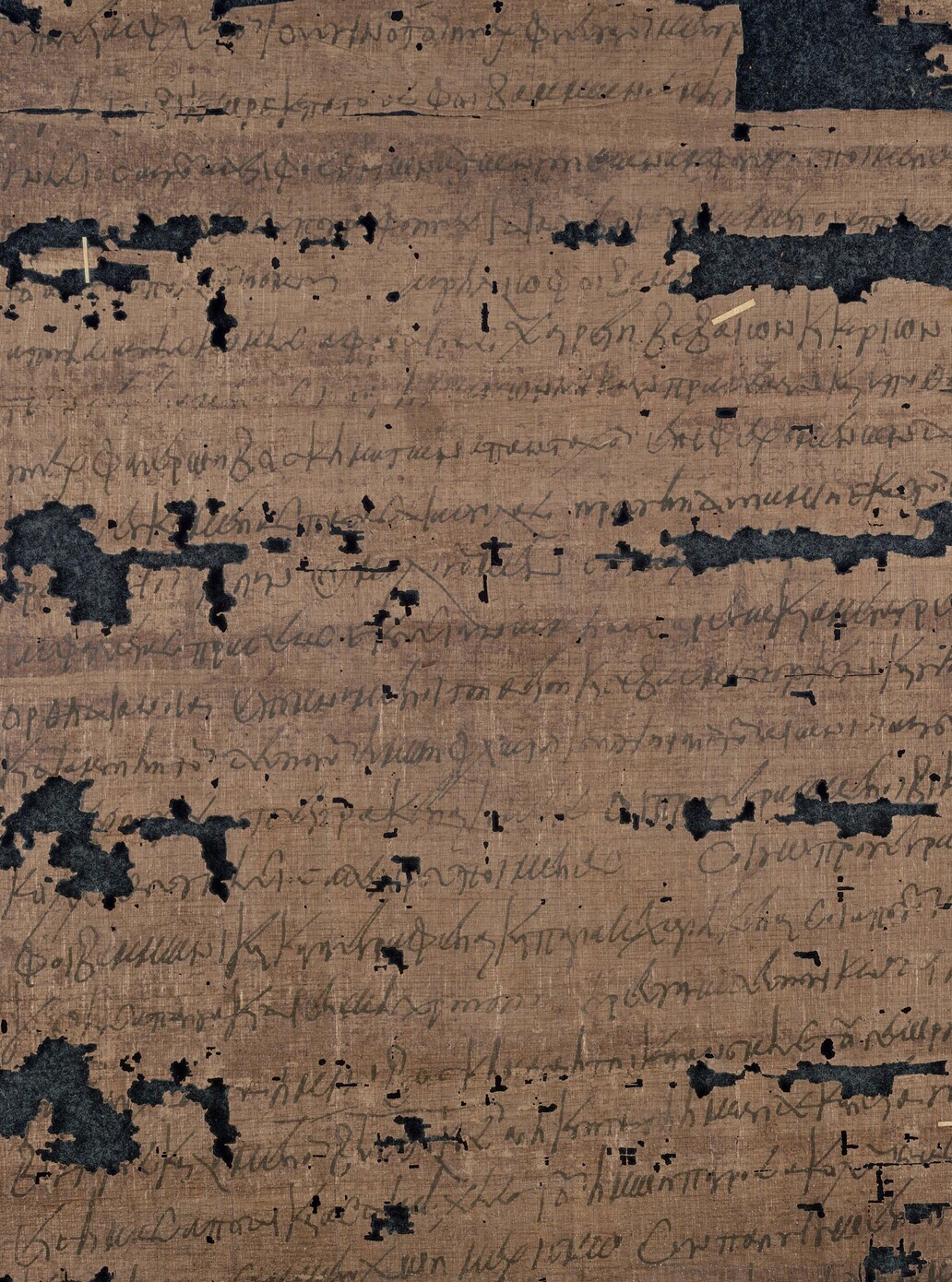}
		};
		\node[right=0.1cm of i1](i2){
			\includegraphics[height=0.3\textheight]{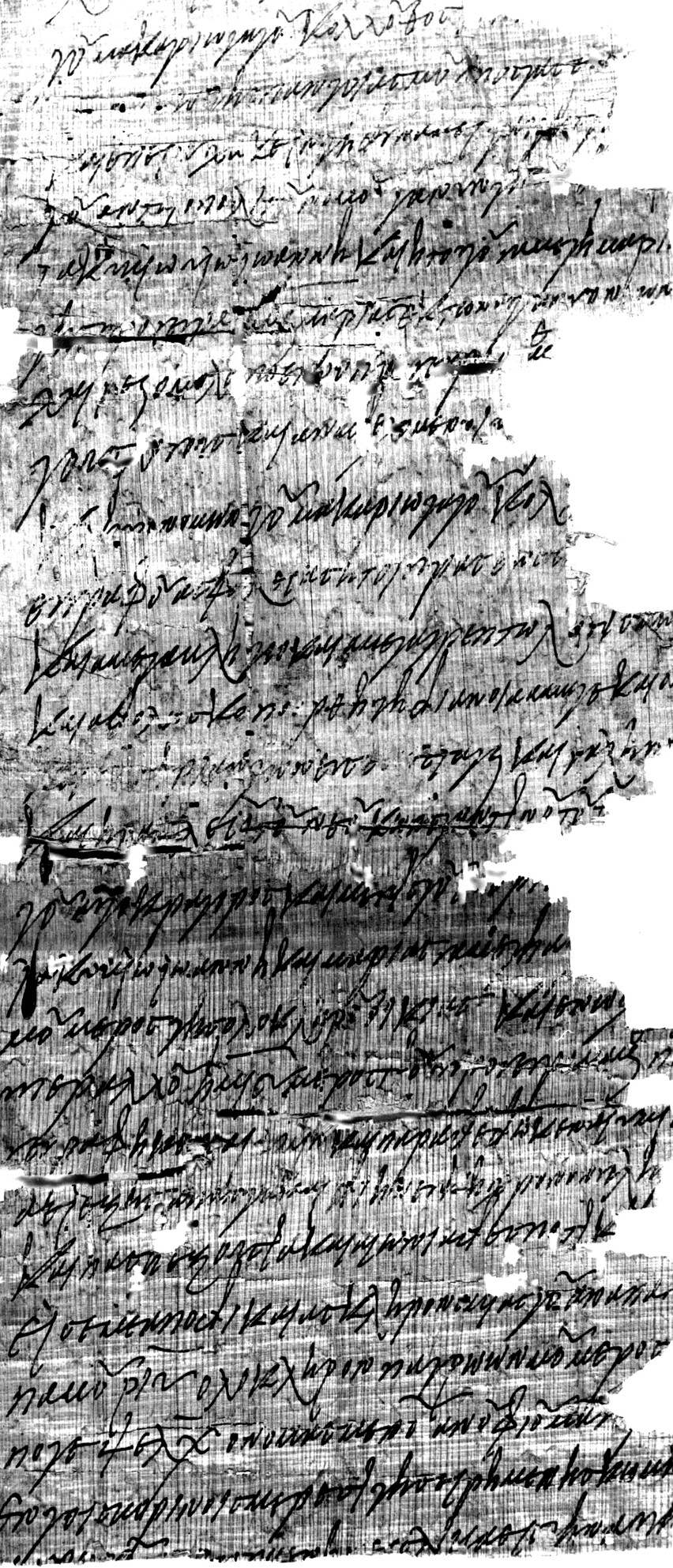}
		};
		\node[right=0.1cm of i2.north east, anchor=north west](i3){
			\includegraphics[height=0.09\textheight]{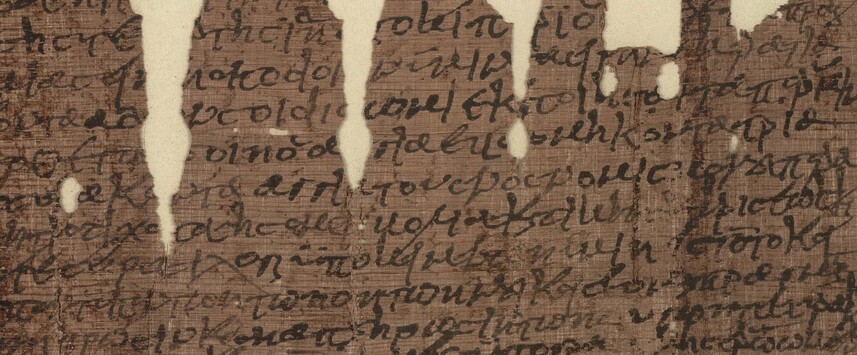}
		};
		\node[anchor=south east] at (i3.east |- i2.south) (i4){
			\includegraphics[height=0.14\textheight]{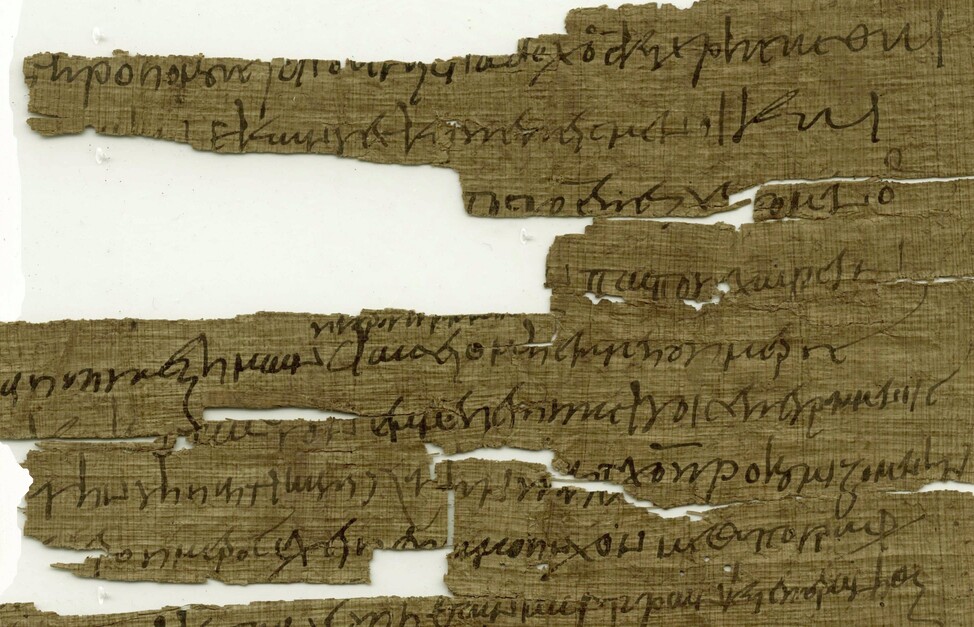}
		};
	\end{tikzpicture}
	\caption{Example images of the GRK-Papyri dataset~\cite{Mohammed19}. 
			IDs: Abraamios\_3, Andreas\_8, Dioscorus\_5, Victor\_2}
	\label{fig:examples}
\end{figure}

\begin{figure}[t]
\centerline{\includegraphics[width=\columnwidth]{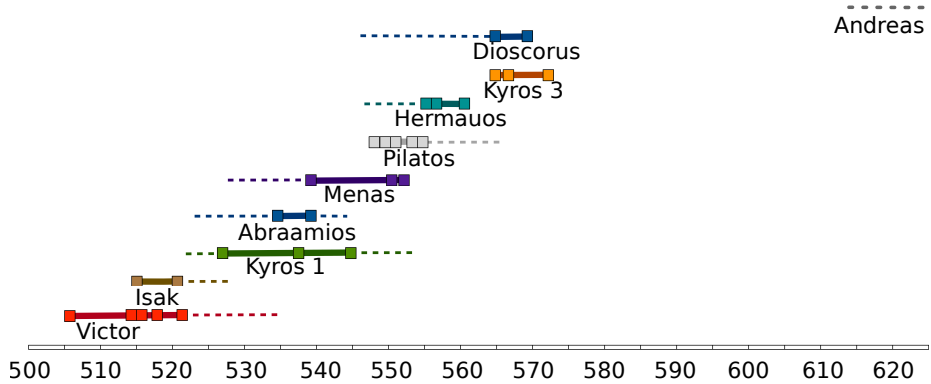}}
\caption{Period of activity of the ten writers. Squares mark the precisely dated texts included in the GRK-Papyri dataset. Dashed lines mark the maximum extension known thanks to other texts outside the dataset. Note that for Andreas, his only text is securely dated from the period marked with a dotted line.}
\label{fig:dates}
\end{figure}

We use the GRK-papyri dataset~\cite{Mohammed19}, see \cref{fig:examples} for some example images. 
It consists of 50 document images written in Greek on papyri by ten different notaries. 
For each writer, four to seven samples were cropped from securely attributed texts: sometimes non-overlapping parts of a single long document, sometimes different documents possibly several years apart from one another. 
They all date from the \nth{6} and early \nth{7} century CE, as can be seen in \cref{fig:dates}. 
The images come from several collections, thus using various digitization parameters.\footnote{Meta-data on the images are available (reference, date, collection...) at \url{https://d-scribes.philhist.unibas.ch/en/gkr-papyri/}.}
The images contain different artifacts, such as holes, low contrast, etc., and are hence difficult to process.

\subsection{Evaluation Protocol}
\label{sec:evaluation_protocol}
We evaluate our method in two ways:
\begin{enumerate}
	\item Leave-one-image-out cross validation. 
		That means, we use each of the 50 images as query image and rank
		the other 49 images according to their similarity with the query image. 
		These results already allow a detailed paleographic interpretation.
		From these ranks, we can compute different metrics: the Top-1,
		Top-5, Top-10 identification rate, which denotes the probability that the
		correct writer is among the first 1, 5, or 10 retrieved, \ie most similar, images (this is also known as \emph{soft} criterion). 
		Additionally, we give the mean average precision (mAP) which takes into account all predictions by taking the mean over all average precisions.  
	\item Classification. For this scenario, we train on 20 samples, two from each writer and test on the remainder. 
\end{enumerate}

\subsection{Experiments}

\begin{figure}[t]
\centering
	\tikzsetnextfilename{sift}
  	\begin{tikzpicture}
		[spy using outlines={rectangle,lens={scale=15},height=4cm,width=7cm, connect spies, ultra thick}]
		\node[](img){
			\includegraphics[height=0.27\textheight]{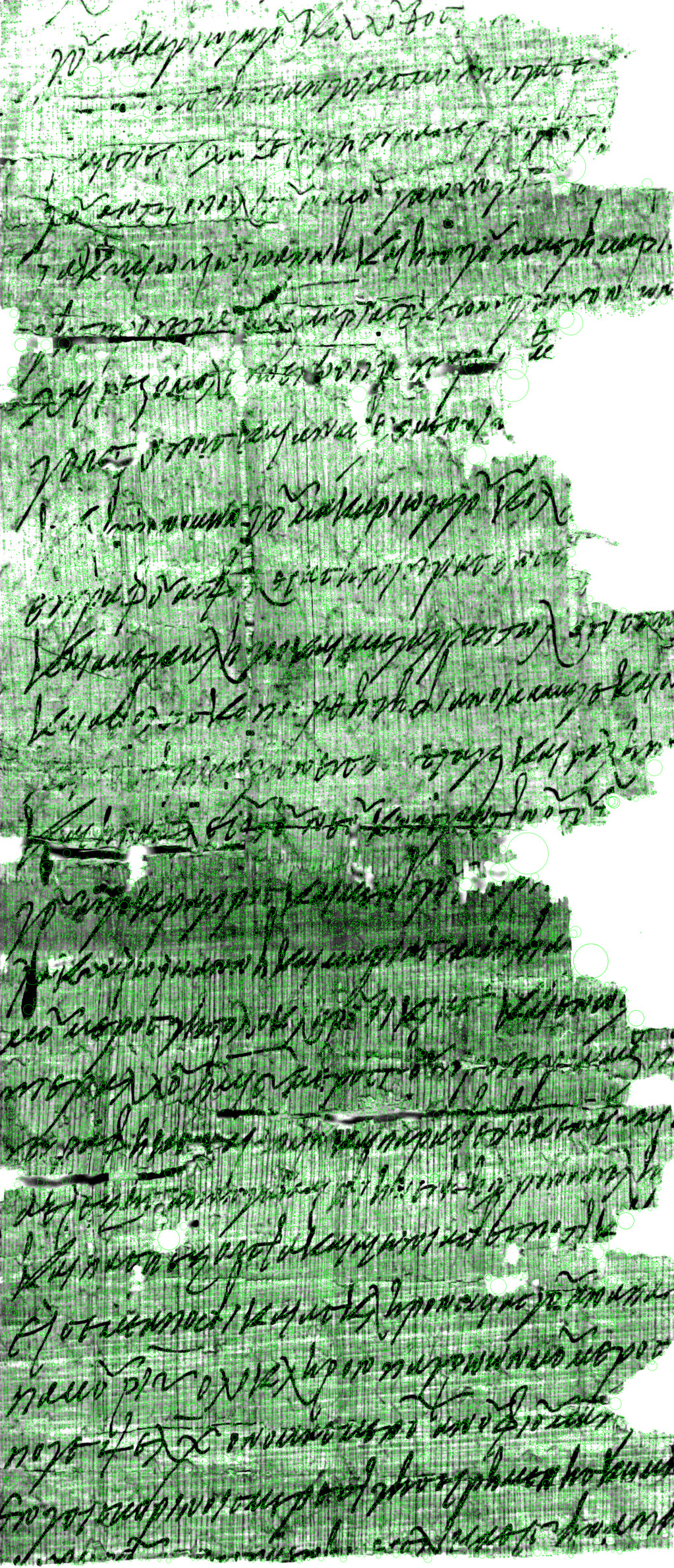}
		};
		\spy[orange,ultra thick] on (0,0) in node [left] at ($(img.east)+(8cm,0)$);
	\end{tikzpicture}
	\caption{Example of \ac{sift} keypoints applied on papyri data. Overlay with GRK-Papyri~\cite{Mohammed19}, ID: Andreas\_8}
  \label{fig:kpts}
\end{figure}
We evaluate five different methods, where the first four methods only differ in the sampling/pre-processing. 
As baseline, we extract the \ac{sift} descriptors at \ac{sift} keypoints.
Keypoints overlaid over one image of the GRK-Papyri dataset can be seen in \cref{fig:kpts}. 
The example shows that many features are lying in the background of the image.

In general, we have two different ways to deal with such a problem: 
\begin{enumerate*}
	\item Make the writer identification method robust against noise (artifacts,
		holes, etc.). This can for example be achieved by using heavy data augmentation
		during the feature learning process.  
	\item Remove the noise by means of \emph{binarization}, \ie a segmentation of
		the writing. 
\end{enumerate*}
We follow the second approach and thus try to segment the text in the papyri data.
\begin{figure}[t]
	\centering
	\tikzsetnextfilename{binarization}
\begin{tikzpicture}
		[spy using outlines={rectangle,lens={scale=3},height=1.8cm,width=3cm, connect spies, ultra thick}]
		\node[](img){
			\includegraphics[height=0.2\textheight]{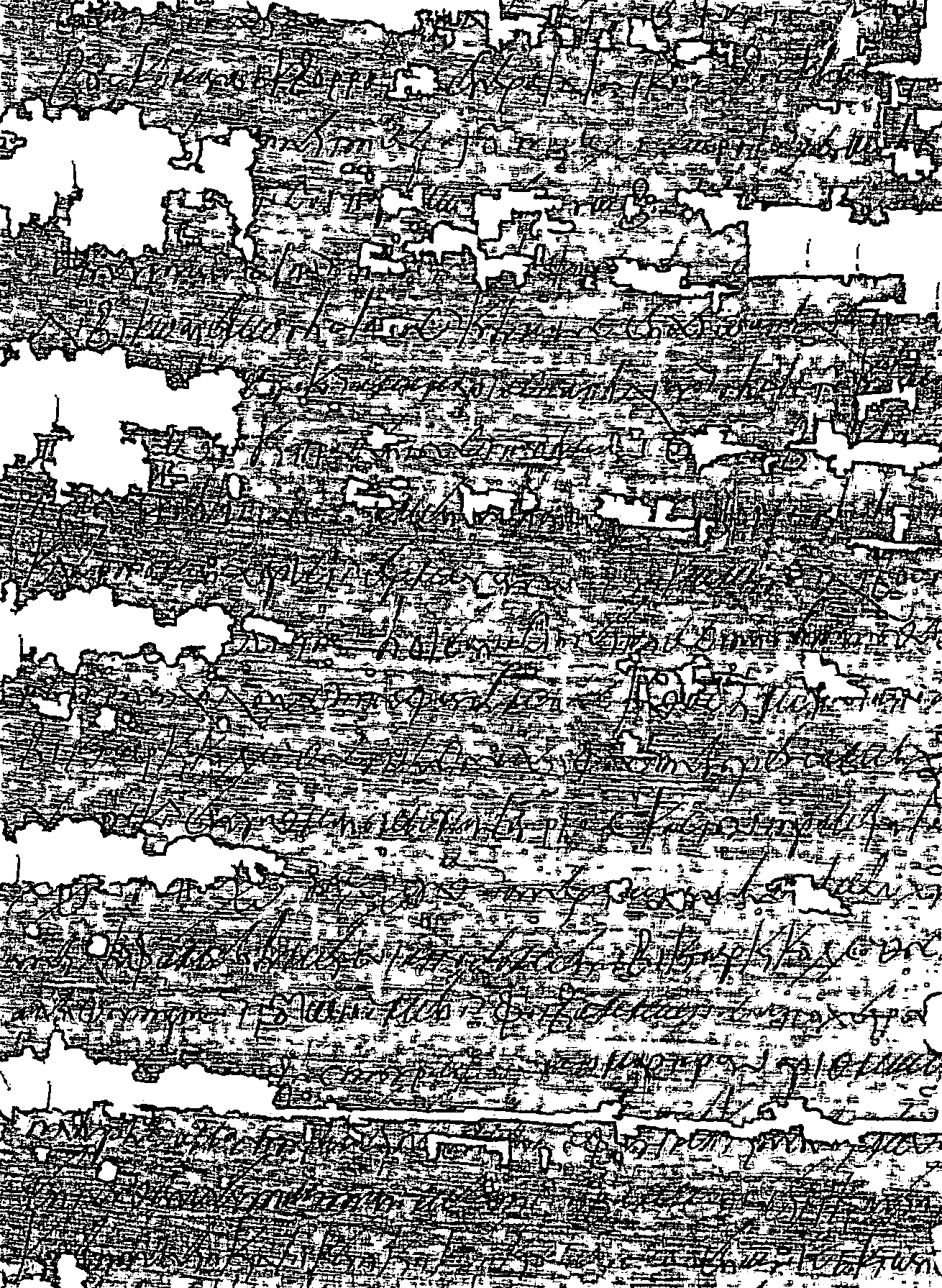}
		};
		\spy[orange,ultra thick] on (0,0) in node [right] at ($(img.east)+(0.1,1)$);

		\node[right=5.5cm of img](i2){
			\includegraphics[height=0.2\textheight]{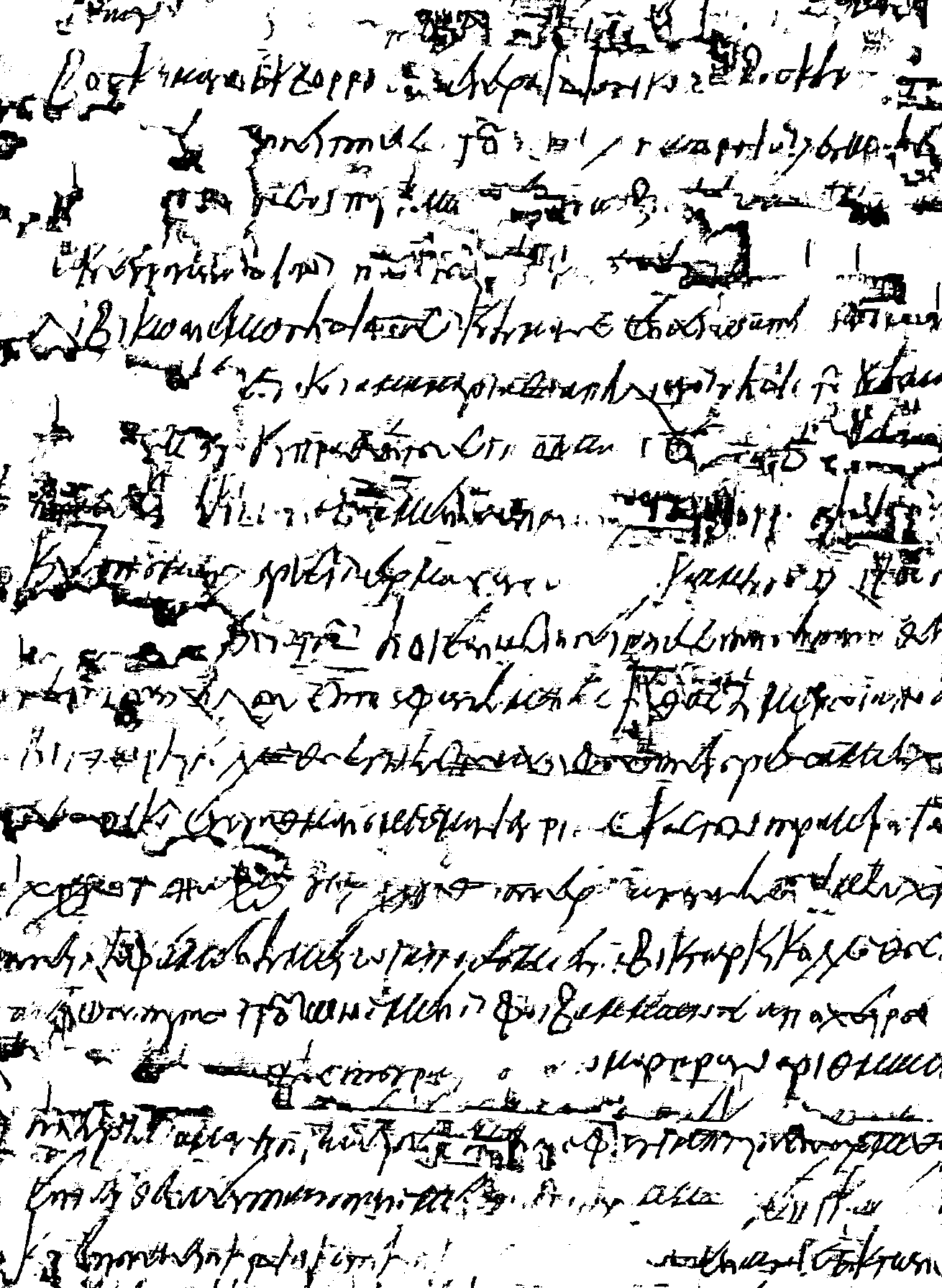}
		};
		\spy[orange,ultra thick] on (i2) in node [left] at ($(i2.west)-(0.1,1)$);

	\end{tikzpicture}
  \caption{Su binarization vs.\ AngU-Net. Image source:
		GRK-Papyri~\cite{Mohammed19}, ID:\ Abraamios\_4.}
  \label{fig:su_vs_angu}
\end{figure}

\subsubsection{Binarization} 
We evaluate two different binarization methods. 
The first one by \etal{Su}~\cite{Su10} works commonly well for such purposes.
However, on some images, the fiber of the papyri data causes sever artifacts,
see for example \cref{fig:su_vs_angu} (left). 

The second binarization method, denoted as \emph{AngU-Net} is based on the popular U-Net~\cite{Ronneberger15}.
The model was trained on \numproduct{512 x 512} patches cropped from the training set of the 2017 DIBCO Dataset~\cite{Pratikakis17}.
The model was specifically trained with augmentations consistent to textual information and designed to simulate material degradation using TorMentor\footnote{\url{https://github.com/anguelos/tormentor}}~\cite{tormentor}.
The effectiveness of this approach is visualized in \cref{fig:su_vs_angu} (right).
For inference, the AngU-Net is used in a fully-convolutional manner. 

\begin{figure}[t]
	\centering
	\tikzsetnextfilename{rsift}
\begin{tikzpicture}
		[spy using outlines={rectangle,lens={scale=3},height=1.8cm,width=3cm, connect spies, ultra thick}]
		\node[](img){
			\includegraphics[height=0.2\textheight]{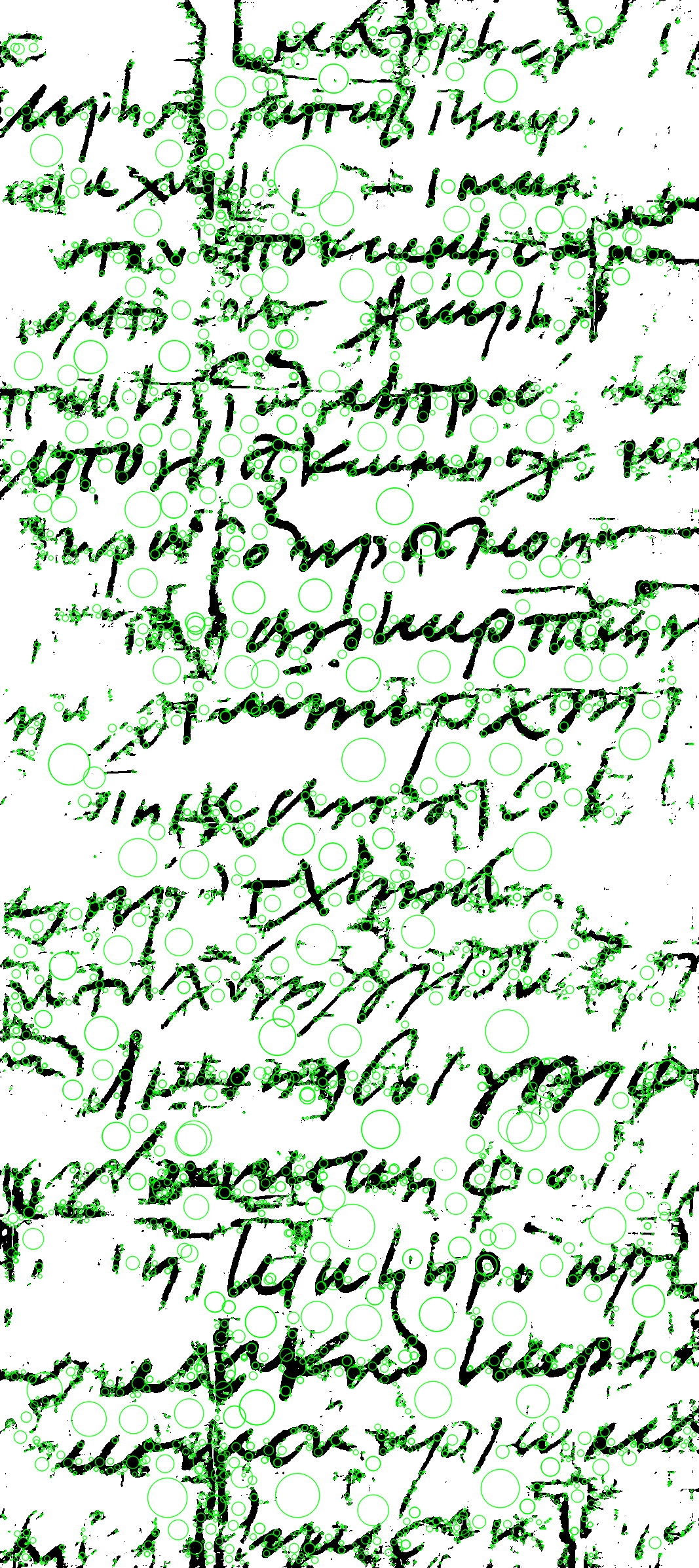}
		};
		\spy[orange,ultra thick] on (0,0) in node [right] at ($(img.east)+(0.1,1)$);

		\node[right=5.5cm of img](i2){
			\includegraphics[height=0.2\textheight]{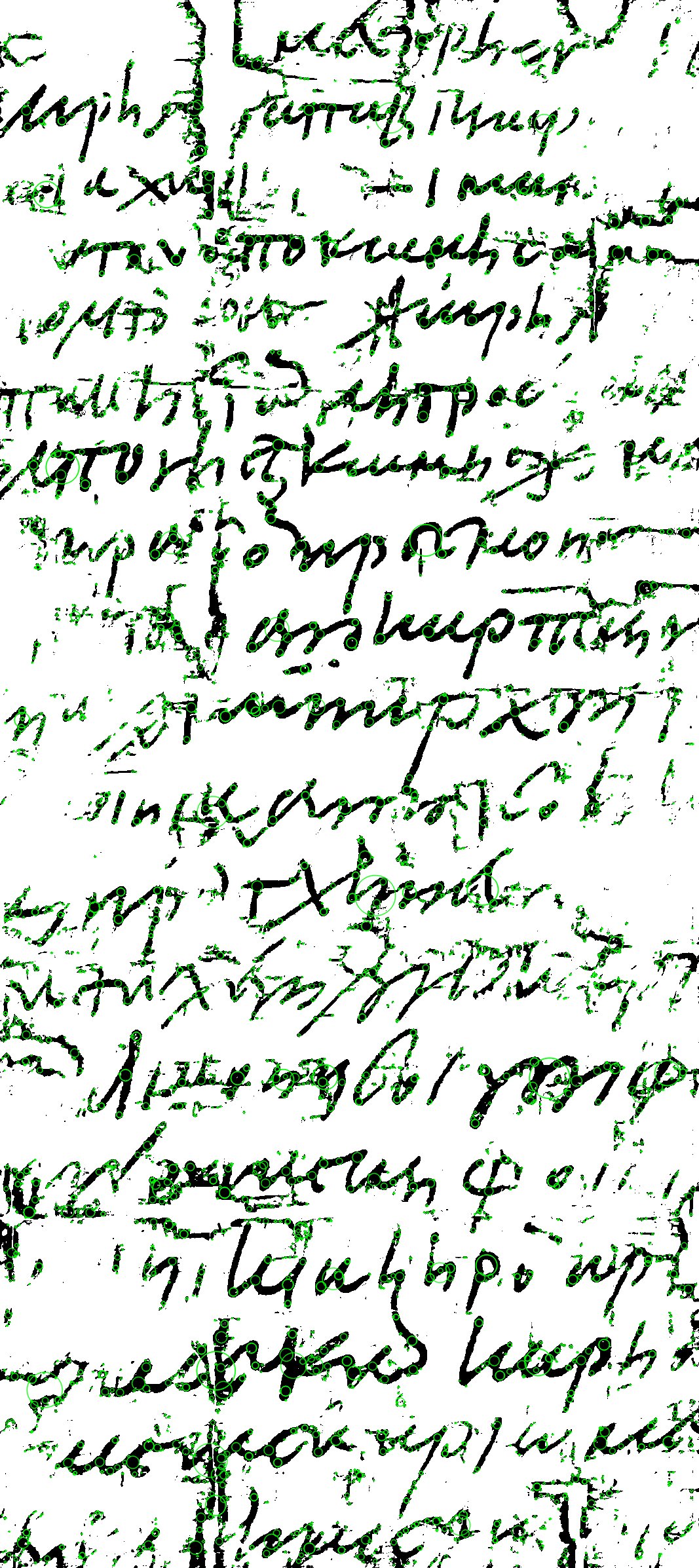}
		};
		\spy[orange,ultra thick] on (i2) in node [left] at ($(i2.west)-(0.1,1)$);

	\end{tikzpicture}
  \caption{Common SIFT keypoints (left) vs.\ restricted SIFT keypoints (right). Image source:
		GRK-Papyri~\cite{Mohammed19}, ID:\ Victor\_5.}
  \label{fig:orig_vs_restricted}
\end{figure}
Given the binarized images, we detect keypoints, extract local features by one of the two presented methods, and compute the global image representations as done in the baseline. 
As an alternative strategy, we can restrict the SIFT keypoints to lie strictly on dark (here:\ black) pixels~\cite{Christlein17ICDAR}.
This can be achieved by restricting the SIFT keypoint extraction such that only minima in the scale space are used.
The effect can be seen in \cref{fig:orig_vs_restricted}. 

\subsubsection{Retrieval Results}
\begin{table}[t]
\centering
	\caption{Writer retrieval results, evaluated by a leave-one-image-out cross-validation.}
	\begin{tabular}{lcccc}
		\toprule
		Method &	Top-1 & Top-5 & Top-10 & mAP\\
		\midrule
		Mohammed et al.~\cite{Mohammed19} & 30 & \\
		\midrule
		SIFT (Baseline) &	28 & 70 & 84 & 30.3\\
		Su Binarization + SIFT & 40 & 72 & 86 & 30.5\\
		AngU-Net + SIFT & 46 & \textbf{84} & 88 & 36.5\\
		AngU-Net + R-SIFT & 48 & \textbf{84} & 92 & \textbf{42.8}\\
		AngU-Net + Cl-S~\cite{Christlein17ICDAR} & \textbf{52} & 82 & \textbf{94} & 42.2\\
		\bottomrule
	\end{tabular}
	\label{tab:retrieval}
\end{table}
We first focus on the first evaluation protocol, \ie image retrieval by applying
a leave-one-image-out cross-validation. 
All results are shown in \cref{tab:retrieval}.
The reference method of \etal{Mohammed}~\cite{Mohammed19}
and our baseline approach achieve quite similar results, where our baseline method is slightly worse, \ie retrieving one less sample, out of the 50, correctly.
However, when we apply binarization, the picture alters drastically. 
The binarization with the method of \etal{Su}~\cite{Su10} gives already
\SI{10}{\percent} better results. 
This can be further improved by using a better segmentation method, \eg using
the proposed AngU-Net. 
Also restricting the keypoints (R-SIFT) to lie on the writing improves the result slightly. 
This is interesting since we encountered the contrary in handwritten Latin text~\cite{Christlein17ICDAR}. 
Finally, the Top-1 accuracy can be further improved by using our self-supervised approach to learn robust local features~\cite{Christlein17ICDAR}. 

\begin{figure*}[t]
  \begin{subfigure}{0.40\textwidth}
    \includegraphics[height=5cm]{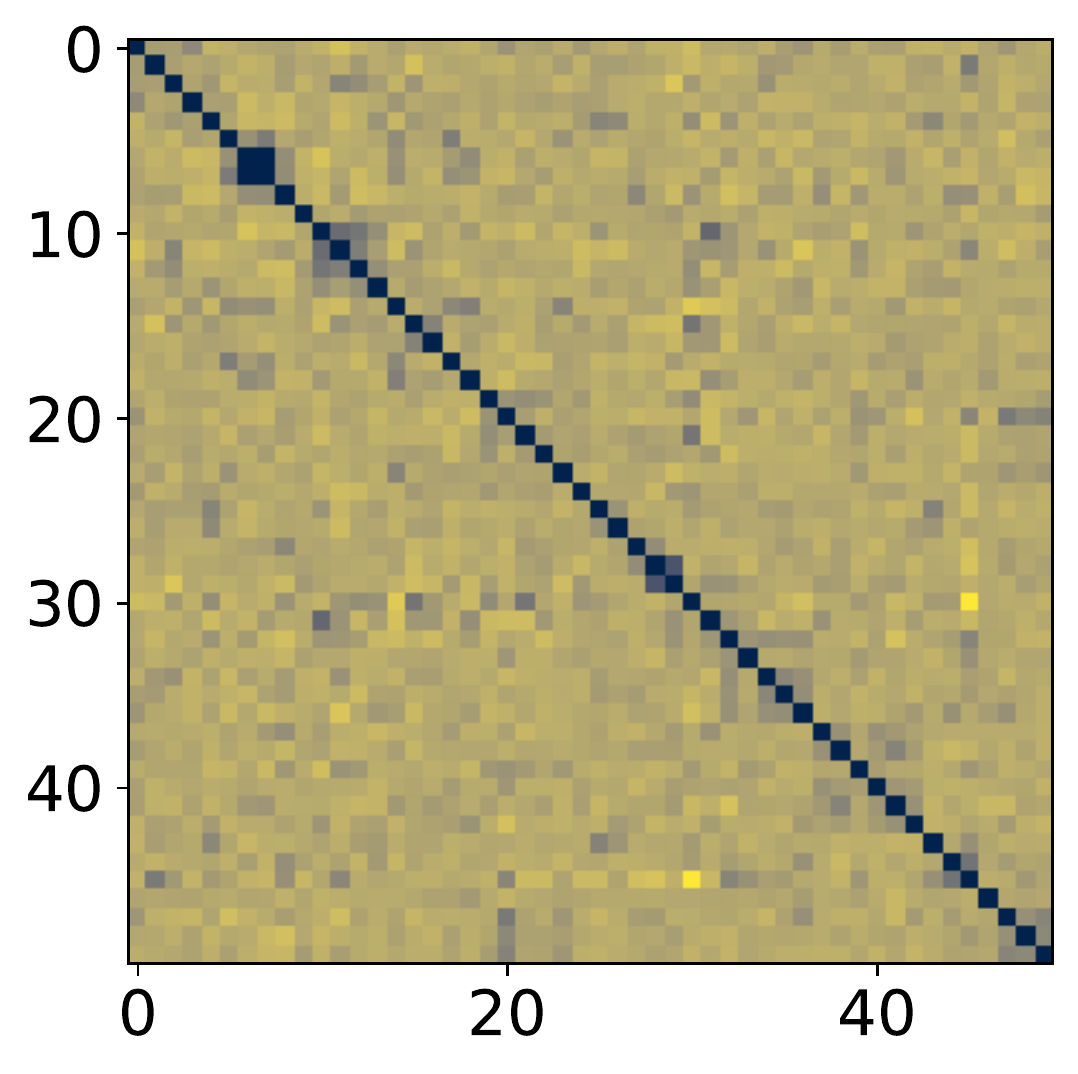}
    \caption{By document}\label{fig:heatmap-a}
  \end{subfigure}%
   \hfill
  \begin{subfigure}{0.58\textwidth}
    \includegraphics[height=5cm]{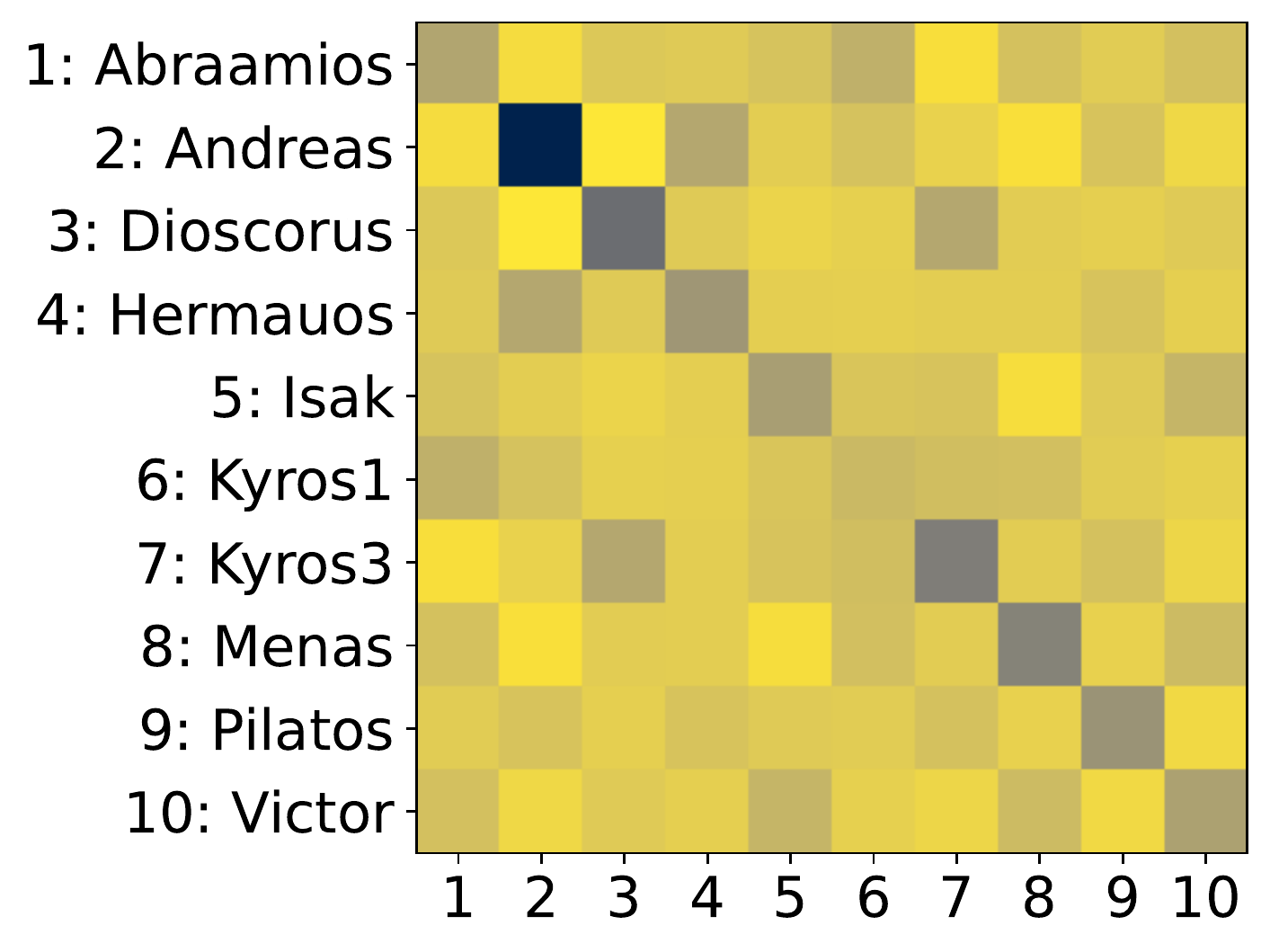}
     \caption{By scribe}\label{fig:heatmap-b}
  \end{subfigure}%
 \caption{Visualization of AngU-Net + R-SIFT method by heatmaps (the darker the square, the higher the similarity).
 In \subref{fig:heatmap-a}, the numbers are related to the writer samples in alphabetic order (from Abraamios\_1 to Victor\_8, Victor \_10 being before Victor\_2).
 In \subref{fig:heatmap-b}, the similarity between scribes is displayed as the average of each pair of images; inter-scribe similarity is computed by ignoring comparisons of images with themselves.
 }
 \label{fig:heatmap}
\end{figure*}

Based on the document-level heatmap, see \cref{fig:heatmap-a}, the best score (smallest number, closest similarity) was achieved between Andreas 5 and 6, \ie two samples from the same document. 
Second best was between Kyros3\_1 and Kyros3\_2, two different texts written only one week apart. 
Among the best scores are Dioscorus 2 and 3 (recto and versos of the same text) as well as Victor 2 and 3, which come from the same document but in two different collections, thus two totally different images.

The scribe-level heatmap, in \cref{fig:heatmap-b}, shows the inter-scribe similarity, \ie the average distance between all documents of a scribe and all documents of another one.
We can see that texts written by Andreas have the highest similarity among them, while the ones of Kyros1 have the lowest. 
The homogeneity of Andreas' samples is not surprising since on the one hand Andreas' hand is represented by four samples of the same document, three coming from the same collection, and on the other hand he is a chronological outsider (by far the most recent writer, see \cref{fig:dates}). 
Kyros1 is represented by four different documents, three of which are precisely dated and span over 18 years, which makes his period of activity the longest present in the dataset, \cf \cref{fig:dates}.
Victor is in a similar situation, since he is represented by 7 samples coming from 6 documents and spanning over 16 years.
However, the similarity between his samples is quite high, suggesting that his handwriting has varied less than Kyros1's one over a comparable amount of time.

\begin{table}[t]
	\centering
		\caption{Classification results.}
\begin{tabular}{lcc}
	\toprule
	Method &	Top-1 & Top-5\\
	\midrule
	Mohammed et al.~\cite{Mohammed19} & 26 & \\ 
	Nasir \& Siddiqi~\cite{Nasir21LFF} & 54 &\\
	Nasir et al.~\cite{Nasir21WCF} & \textbf{64} &\\
	\midrule
	AngU-Net + \phantom{R-}SIFT + NN & 47 & 83\\
	AngU-Net + \phantom{R-}SIFT + SVM & 57 & \textbf{87}\\
	AngU-Net + R-SIFT + NN & 53 & 77\\
	AngU-Net + R-SIFT + SVM & \underline{60} & 80\\
	\bottomrule
	\end{tabular}
	\label{tab:classification}
\end{table}

\begin{table}[t]	
	\centering
	\caption{Confusion matrix of the classification result obtained by  using AngU-Net + R-SIFT + SVM. Correct ones highlighted in blue, wrong ones in red.}
\newcommand\items{10}   %
\arrayrulecolor{white} %
\noindent\begin{tabular}{lrc*{\items}{|c}|}
& \raisebox{1em}{Predicted$\rightarrow$} &
\multicolumn{1}{c}{\rot{Abraamios}} & 
\multicolumn{1}{c}{\rot{Andreas}} & 
\multicolumn{1}{c}{\rot{Dioscorus}} & 
\multicolumn{1}{c}{\rot{Hermauos}} & 
\multicolumn{1}{c}{\rot{Isak}} & 
\multicolumn{1}{c}{\rot{Kyros1}} & 
\multicolumn{1}{c}{\rot{Kyros3}} & 
\multicolumn{1}{c}{\rot{Menas}} & 
\multicolumn{1}{c}{\rot{Pilatos}} & 
\multicolumn{1}{c}{\rot{Victor}} \\

\multirow{10}{*}{\rotn{True Label}} 
&Abraamios 	& \cellcolor{blue!20}1&  0&  0&  0&  0&  \cellcolor{red!20}1&  0&  0&  0&  0\\ \hhline{~*\items{|-}|} 
&Andreas 		& 0&  \cellcolor{blue!20}1&  0&  \cellcolor{red!20}1&  0&  0&  0&  0&  0&  0\\ \hhline{~*\items{|-}|}
&Dioscorus 	& 0&  0&  \cellcolor{blue!20}2&  \cellcolor{red!20}1&  0&  0&  0&  0&  0&  0\\ \hhline{~*\items{|-}|}
&Hermauos 	& 0&  0&  0&  \cellcolor{blue!20}1&  0&  \cellcolor{red!20}2&  0&  0&  0&  0\\ \hhline{~*\items{|-}|}
&Isak 			& 0&  0&  0&  \cellcolor{red!20}1&  \cellcolor{blue!20}1&  0&  0&  0&  0&  \cellcolor{red!20}1\\ \hhline{~*\items{|-}|}
&Kyros1 		& 0&  \cellcolor{red!20}1&  0&  0&  0&  \cellcolor{blue!20}1&  0&  0&  0&  0\\ \hhline{~*\items{|-}|}
&Kyros3 		& 0&  0&  0&  0&  0&  \cellcolor{red!20}1&  \cellcolor{blue!20}1&  0&  0&  0\\ \hhline{~*\items{|-}|}
&Menas 			& 0&  0&  0&  0&  0&  0&  0&  \cellcolor{blue!20}3&  0&  0\\ \hhline{~*\items{|-}|}
&Pilatos 		& 0&  0&  0&  0&  \cellcolor{red!20}1&  0&  0&  0&  \cellcolor{blue!20}3&  0\\ \hhline{~*\items{|-}|}
&Victor 		& 0&  0&  0&  0&  0&  \cellcolor{red!20}1&  0&  \cellcolor{red!20}1&  0&  \cellcolor{blue!20}3\\ \hhline{~*\items{|-}|}
\end{tabular}	

\label{tab:confusion}
\end{table}

\subsubsection{Classification Results}
Finally, we conduct the second evaluation protocol, \ie we have a train/test split and use two samples of each writer for training. 
We evaluate two different classifiers: a Nearest Neighbor (NN) approach and a \ac{svm}.
For the latter approach, we train for each writer an individual \ac{svm} using the two samples as positives and the remaining 18 samples as negatives.
The two classes are balanced by weighting them indirectly proportionally to the number of respective samples. 
The classifiers use the global representations computed from local descriptors sampled on normal SIFT keypoints or restricted keypoints (R-SIFT).

The results in \cref{tab:classification} reflect the same benefit of proper binarization as in the retrieval case. 
Comparing SIFT and R-SIFT, the latter is beneficial also for classification purposes. 
Classifier-wise, the use of SVMs is preferable in comparison to a simple nearest neighbor classifier, although the SVMs were only trained with two positive samples. 
Since the writer classification is fully supervised, we refrain from evaluating the unsupervised CL-S method, which would be needed to be trained on an even smaller training set (30 instead of 50 images) in an unsupervised manner. 
In future work, it might be worth investigating if an unsupervised pre-training on a large papyri corpus instead of using an ImageNet-pretrained network is beneficial for an additional fine-tuning step similar to the methods proposed by \etal{Nasir}~\cite{Nasir21LFF,Nasir21WCF}.
The full confusion matrix can be seen in \cref{tab:confusion}. 
Interestingly, the writer who was the least confused with anybody else is the geographic outsider Menas. Indeed, while all the others lived and worked in the same village, Menas' dossier comes from a big city several hundred kilometers north.   
Another interesting result, which requires an in-depth paleographic analysis, is the confusion between Hermauos and Kyros1 as well as the many false predictions they both have generated. 
We have already mentioned the variety among Kyros1's samples.
Hermauos' results may be due to the important degradation of some of his samples. 
Future investigations will aim determining if these two writers share specific paleographic features.

In comparison with the state of the art~\cite{Nasir21WCF}, the proposed \ac{sift}-based method does not fully compete to the deep learning based methods that train a \ac{cnn} on small image patches and then apply a majority vote. 
While the state-of-the-art method~\cite{Nasir21LFF} is slightly superior, a drawback of it is that it cannot be used for novel writers, the \ac{cnn} is tuned towards the writers of the training set and needs to be fine-tuned for each novel writer. 
In contrast, our method can easily be adapted to more writers by computing a new \ac{svm} or just by means of nearest neighbor matching. 

\section{Conclusion}\label{sec:conclusion}
In this work, we investigated automatic writer identification on the specific historical documents that are the Greek papyri. 
In particular, we evaluated different binarization and sampling procedures in two different scenarios:\ retrieval and classification.
We show that binarization, especially deep learning-based binarization, improves the writer identification performance by removing the most noise and artifacts introduced by the papyri. 
We believe that better binarization methods can help to reduce the misclassifications further. 
Additionally, a sampling that is restricted to the handwriting is beneficial.
The obtained results are already stimulating for further paleographic investigations.
Some expected results have been confirmed: the geographical and chronological outsiders have distinguished themselves. Some have been refuted: Abraamios was supposed to have a particularly clumsy hand, easy to recognize, but this has not been the case in neither of the two scenarios.  
For future work, we would like to investigate the possibility of improving writer identification by learning noise-robust descriptors. 
We would also lead in-depth interpretations of the results in a paleographic perspective to better apprehend and qualify the similarities or at the opposite the originality of the various hands, setting the foundations for sounder scribal attributions on paleographic and computer-assisted grounds.   

\section*{Acknowledgement}
This work was partially supported by the Swiss National Science Foundation as part of the project no.\ PZ00P1-174149 ``Reuniting fragments, identifying scribes and characterizing scripts: the Digital paleography of Greek and Coptic papyri (d-scribes)''.
This research was supported by grants from NVIDIA and utilized NVIDIA Quadro RTX 6000. %
\bibliographystyle{splncs04}
\bibliography{references}

\end{document}